%% file: paper.tex
\documentclass[runningheads]{llncs}
\input{aux/packages.tex}

\usepackage{xcolor}         %

\begin{document}
\title{Tertiary Lymphoid Structures Generation through Graph-based Diffusion} %
\author{Manuel Madeira\inst{1}\orcidID{0000-0002-8205-404X} \and
Dorina Thanou\inst{1}\orcidID{0000-0003-2319-4832} \and
Pascal Frossard\inst{1}\orcidID{0000-0002-4010-714X}}

\authorrunning{M. Madeira et al.}
\institute{\'Ecole Polytechnique F\'ed\'erale de Lausanne (EPFL), Lausanne, Switzerland \\
\email{manuel.madeira@epfl.ch}\\
}

\maketitle              %
\input{content/abstract}
\input{content/introduction.tex}

\input{content/background.tex}

\input{content/method.tex}

\input{content/experiments.tex}
\input{content/conclusion.tex}
\input{content/acknowledgements}
\clearpage
\bibliographystyle{splncs04}
\bibliography{aux/new_bibliography.bib}

\clearpage
\input{content/appendix}

\end{document}

%% file: aux/packages.tex
\usepackage{graphics}
\usepackage{hyperref}
\usepackage{multirow}
\usepackage{mathtools, amssymb}
\usepackage{cleveref}
\newcommand{\ka}[1]{\kappa(#1)}
\usepackage{graphicx}
\usepackage{booktabs}

\usepackage[T1]{fontenc}
\usepackage{graphicx}
\usepackage{color}

%% file: content/abstract.tex
\begin{abstract}
Graph-based representation approaches have been proven to be successful in the analysis of biomedical data, due to their capability of capturing intricate dependencies between biological entities, such as the spatial organization of different cell types in a tumor tissue.
However, to further enhance our understanding of the underlying governing biological mechanisms, it is important to accurately capture the actual distributions of such complex data.
Graph-based deep generative models are specifically tailored to accomplish that.
In this work, we leverage state-of-the-art graph-based diffusion models to generate biologically meaningful cell-graphs. 
In particular, we show that the adopted graph diffusion model is able to accurately learn the distribution of cells in terms of their tertiary lymphoid structures (TLS) content, a well-established biomarker for evaluating the cancer progression in oncology research.
Additionally, we further illustrate the utility of the learned generative models for data augmentation in a TLS classification task. To the best of our knowledge, this is the first work that leverages the power of graph diffusion models in generating meaningful biological cell structures.

\keywords{Deep generative models \and Graph-based diffusion \and Tertiary lymphoid structures}
\end{abstract}

%% file: content/introduction.tex
\section{Introduction}

Biomedical applications have largely benefited from graph-based approaches as a powerful framework for complex interactions modelling. The representation and analysis of the relationships between biologically-relevant entities has been exploited at different levels, ranging from more abstract dependencies, such as metabolic or gene regulatory networks~\cite{van2002graph,de2002modeling,li2022graph}, to completely observable relations, such as the spatial interactions of cells in digital pathology (DP) settings~\cite{jaume2021quantifying,jaume2021histocartography,ahmedt2022survey}.

Under these settings, we are typically interested in tasks such as the understanding of metabolic or genetic interactions, patient diagnosis, or response prediction. A key factor to perform satisfactorily on those problems is the inference over previously unseen graphs and, to accomplish so, we need to correctly capture the underlying governing biological mechanisms.
In particular, graph generation models seek to model the distribution of the actual graphs they are trained on. Thus, this type of approach can play a pivotal role for a better understanding of graph distributions, as well as on the generation of biologically plausible graph instances. 

Diffusion models~\cite{sohl2015deep,ho2020denoising} have recently emerged as the state-of-the-art approach for deep generative modelling by combining several desirable properties such as training stability, excellent sample quality, easy model scaling, and good distribution coverage. Meanwhile, several graph-based diffusion schemes have been proposed through the adaptation of either score-based generative modelling~\cite{niu2020permutation,jo2022score} or of discrete denoising diffusion probabilistic models~\cite{austin2021structured,vignac2022digress,haefeli2022diffusion} to graphs. Adapting these to biological applications amenable to graph-based modelling holds great promise for further improvement in biological graphs mimicking.

\begin{figure}[t!]
    \includegraphics[width=\textwidth]{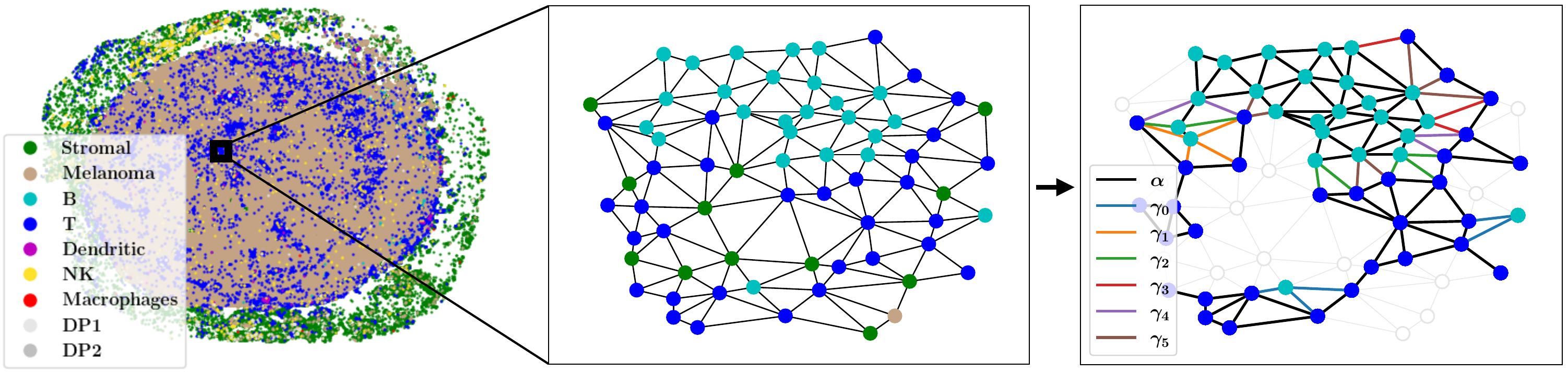}
    \centering
        \caption{Cell-graphs extracted from a global cell-graph, computed from a cancerous tissue sample imaged through multiplexed immunofluorescence with several cell type markers. \emph{Left:} Cell-graph obtained from a whole-slide image. \emph{Center:} High TLS content subgraph extracted from the cell-graph on the left. It is distinctly composed of a B-cell nucleus enveloped by T-cells. \emph{Right:} Subgraph used for the TLS embeddings computation, following the procedure described in \cite{schaadt2020graph}. Only the edges between B- and T-cells are kept. If the vertices of the edge are of the same cell type, it is classified as $\alpha$; otherwise, as $\gamma_k$, with $k$ consisting of the number of B-cells neighboring the B-cell that is a vertex of the edge.}
    \label{fig:cell_graph}
\end{figure}

In this work, we leverage on previously proposed graph-based diffusion approaches and extend their application to the cell-graph representation of biological tissues~\cite{gunduz2004cell,jaume2021histocartography}. Such graphs consider physical proximity between cells in tissue slides as a proxy to their interaction, assigning cells to nodes and drawing edges between adjacent cells, as depicted in \Cref{fig:cell_graph}. %
We focus specifically on the simple, yet biologically meaningful, tertiary lymphoid structures (TLS). TLS are highly structured biological entities composed of B-cell clusters surrounded by supporting T-cells, typically found in ectopic sites of chronic inflammation~\cite{pitzalis2014ectopic,schaadt2020graph}. These structures are correlated to a longer disease-free survival in cancer~\cite{helmink2020b,lee2016tertiary,dieu2016tertiary,munoz2020tertiary,schaadt2020graph}.
We then build on DiGress~\cite{vignac2022digress}, a state-of-the-art graph generative model for the setting at hand (i.e., small graphs with categorical node and edge features), and extend it to the generation of cell-graphs with high and low TLS content. 

We show that the trained generative model can indeed capture the underlying distribution by evaluating the similarity between the training graph population and a generated graph population in terms of TLS content~\cite{schaadt2020graph}. To further illustrate the pertinence of the model to address the complexity of this task, we compare its performance to a non deep learning graph-based baseline.
Furthermore, we also showcase the utility of generative models for data augmentation in a scarce data regime. We consider a binary classification task, where a Graph Convolutional Network (GCN) is used to classify input graphs as having high or low TLS content. We use the synthetically generated data to augment the training set of the GCN. We demonstrate that the generated graphs are sufficiently faithful to the real data distribution, leading to improved performance on the downstream classification task.

To the best of our knowledge, this work consists of the first graph-based generative approach for cell-graphs. It opens new promising venues in modelling distributions of structured tissues at a cell level, which can be a significant first step towards building more effective machine learning methods in digital pathology or personalised oncology.

%% file: content/background.tex
\section{Background}

In this section, we start by introducing how graphs can be used to model biological entities and their relations. Then, we describe the relevance of generative models in the biomedical realm. We also briefly review graph-based diffusion methods. 

\paragraph{Modelling biomedical structures with graphs.}
In DP, tissue slides are digitized into whole-slide images (WSIs). The predominant applications of deep learning in DP rely on the extraction of image-level representations from those WSIs to a wide variety of tasks, ranging from object recognition, such as slide segmentation or structure detection, to higher level problems, such as cancer grading or survival prediction~\cite{bera2019artificial,serag2019translational}. However, these approaches are typically limited in several aspects. WSIs are typically huge, requiring the tiling of the original image into smaller sized patches to be fed into an image based deep learning approach. This limitation imposes a trade-off on the context per patch that can be considered. Moreover, image based deep learning approaches model pixelwise relations, thus lacking efficient representations of biological entities and their relations. This fact also leads to a more convoluted interpretability of the obtained models. In contrast, entity-graphs based approaches have been shown to evade these limitations~\cite{jaume2021histocartography,ahmedt2022survey}, yielding promising results both in predictive performance and interpretability~\cite{wu2022graph}. 
Such graphs are composed of biological entities as nodes and relations between entities as edges~\cite{gunduz2004cell,jaume2021histocartography}, directly operating at a biologically meaningful level.
As a consequence, representations of tissue structures have shown enhanced explainability~\cite{jaume2021quantifying} and enable the direct interpretation by domain experts. Motivated by these premises, throughout this paper we focus on cell-graphs, assigning cells to nodes and setting edges between adjacent cells.

\paragraph{Generative AI for biomedicine.}

In several biomedical settings, further application and development of data-intensive deep learning methods has been hindered mainly by a lack of high-quality annotated samples, as well as by extensive ethical and privacy regulations. Such is the case for DP and the usage of synthetic data has emerged as a promising research direction to address its scarcity of WSIs. Moreover, the pre-processing steps of obtaining entity-graphs from WSIs, such as stain normalization, image segmentation, or entity detection, lack a standardized framework, being cumbersome, time-consuming and handicapping reproducibility~\cite{jaume2021histocartography}. The possibility of directly generating synthetic entity-graphs jointly tackles those limitations. 

Several ways to augment medical data have been explored, ranging from adding slightly but naively transformed copies of already existing data~\cite{chlap2021review}, which often lead to unexpected distribution shifts, to entire physical simulations~\cite{tang2021augmenting}, which impose a heavy computational burden. Currently, the most promising direction is the use of deep generative models, which seek to generate samples following the same data distribution as their training dataset leveraging on deep learning models. In particular, diffusion models~\cite{sohl2015deep,ho2020denoising} emerged as the state-of-the-art approach for deep generative modelling by combining several desirable properties such as training stability, excellent sample quality, easy model scaling, and good distribution coverage. In the specific context of biological structure modelling, these have been applied to tissue slide images~\cite{moghadam2023morphology}, outperforming the previously predominant Generative Adversarial Networks~\cite{jose2021generative}. Nevertheless, all the existing approaches have remained at the pixel level, aiming to directly generate new DP images and, thus, suffering of the aforementioned image-based deep learning limitations. By deploying diffusion models at the graph level, we seek to combine the generative potential of the former with the efficient modelling capabilities of relation-aware representations of the latter.

Recently, several graph-based diffusion schemes have been proposed through the adaptation of either score-based generative modelling~\cite{niu2020permutation,jo2022score} or of discrete denoising diffusion probabilistic models (D3PMs)~\cite{austin2021structured} adapted to graphs~\cite{vignac2022digress,haefeli2022diffusion}. For the first class of methods, graphs are embedded in a continuous space and Gaussian noise is used to corrupt its node features and adjacency matrix. This approach neither preserves the inherent sparsity of graphs nor scales for larger instances. Conversely, DiGress, a method of the second class, performs diffusion on fully discrete data structures~\cite{vignac2022digress}. Moreover, it also promotes the sparsity of the generated graphs by choosing an adequate noise model, and enhances the expressivity of the used denoising graph neural network (GNN) by adding extra features to its input, leading to state-of-the-art results. For these reasons, DiGress is the graph-based diffusion method that we adopt in this paper.

%% file: content/method.tex
\section{Diffusion based cell-graph generation of Tertiary Lymphoid Structures} %

In this section, we start by framing the problem at hand in more detail and then introduce the graph diffusion method adopted in this work.

\paragraph{Problem formulation.}

We build a cell-graph by assigning cells as nodes and by setting edges between neighboring cells to represent a proxy for spatial cell-cell interactions, which are solely local. Node features represent cell types
that we assume to biologically characterize a cell both anatomically and physiologically.
More formally, provided a graph $G$, we denote by $x_i$ the node attribute (cell type) of node $i$ and by $\textbf{x}_i \in \left\{0,1\right\}^b$ its one-hot encoding, as we consider a total of $b$ different cell types. These are stored in a matrix $\mathbf{X} \in \left\{0,1\right\}^{n \times b}$. Similarly, we denote by $e_{ij}$ the edge attribute between the $i$-th and $j$-th nodes and by $\textbf{e}_{ij}$ its one-hot encoding, that is stored in $\textbf{E}$. As the generative model treats the absence of an edge as a specific edge type, we have $\textbf{e}_{ij} \in \left\{0,1\right\}^{c+1}$, where $c$ is the number of edge types considered and, consequently, $\textbf{E} \in \left\{0,1\right\}^{n \times n \times (c+1)}$. Thus, we unequivocally consider cell-graphs in the form of an unweighted graph $G = (\mathbf{X}, \mathbf{E})$.

In these cell-graphs, we are mostly interested in informative biological structures, such as the TLSs~\cite{helmink2020b,lee2016tertiary,dieu2016tertiary,munoz2020tertiary,schaadt2020graph}. These are composed of clusters of B-cells supported by T-cell compartments and have been shown to be meaningful for medical prognosis in cancer. Their presence in cell-graphs can be measured through the TLS-like organization metric, $\ka{a}$, allowing the computation of the TLS content given a cell-graph~\cite{schaadt2020graph}. 
This computation only considers the edges between B- and T-cells: edges between two cells of the same type are considered a type $\alpha$ edge, while edges between B- and T-cells are considered a type $\gamma_k$ edge, where $k$ is the number of B-cells adjacent to the vertex B-cell. The classification of different edges of a cell-graph into these classes is illustrated in \Cref{fig:cell_graph}.
Provided a cell-graph, its $\ka{a}$ value is defined as the proportion of its $\gamma$ edges whose index is larger than $a$ and, consequently, monotonically decreasing with $a$: 
\begin{equation}\label{eq:tls_embedding}
    \ka{a} = \frac{|E| - |E_\alpha| - \sum^{a}_{k=0} |E_{\gamma_k}|}{|E| - |E_\alpha|},
\end{equation}
where $|E|$, $|E_\alpha|$, and $|E_{\gamma_k}|$ respectively denote the number of edges, of $\alpha$ edges and of $\gamma_k$ edges in the given graph~\cite{schaadt2020graph}.

Here, we consider $a \in \{0,\ldots,5\}$, since we verified empirically that $\ka{a}$ for larger $a$ is non-informative (mostly 0 or, in the rare exceptions, extremely close to that value; check \Cref{app:dist_embeddings} for further details). Consequently, each cell-graph is characterized by its \emph{TLS embedding}, $[\ka{0}, \ldots, \ka{5}] \in \mathbb{R}^6$. Using the TLS embedding of cell-graphs, we are able to characterize distributions of such data structures. Based on insights from the paper proposing the TLS embedding~\cite{schaadt2020graph} and from empirical validation, we can identify two distinct populations of cell-graphs: high TLS content cell-graphs, characterized by $\ka{2}>0.05$ and, conversely, low TLS content cell-graphs, which verify $\ka{1}<0.05$. 

The problem that we consider here is the generation of synthetic data that would be representative of each of the two populations of cell-graphs. Namely, we are interested in building a generative model that can be trained on a few actual samples that are representative of low and high TLS content cell-graphs, respectively, and that can eventually generate new cell-graphs following the same distribution of those two populations.

\begin{figure}[t!]
    \includegraphics[width=\textwidth]{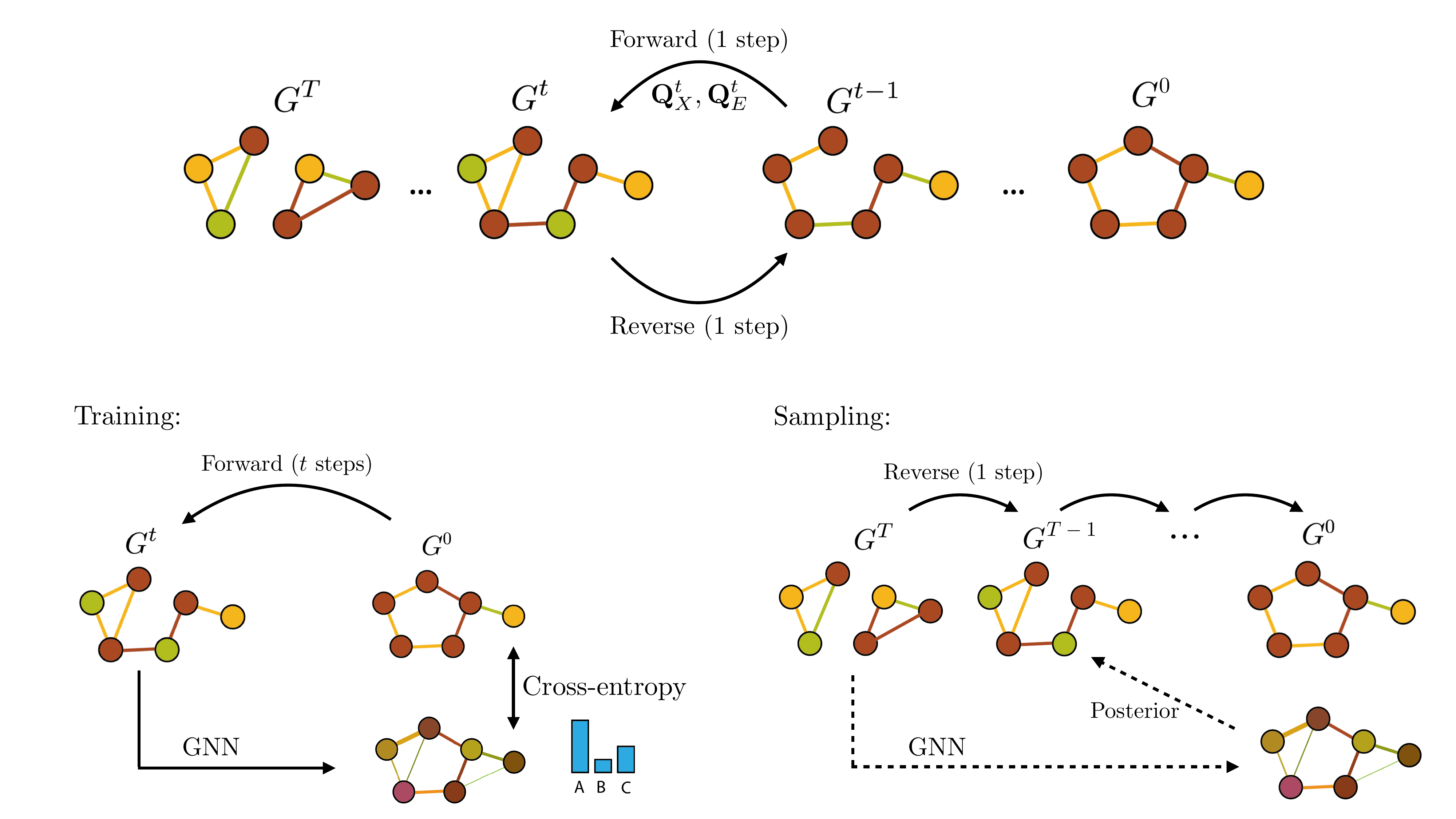}
    \centering
        \caption{DiGress is composed of a forward process, whose noise model is modulated by $\mathbf{Q}_E^t$ and $\mathbf{Q}_X^t$, and a reverse process. In the training process, a GNN is trained to predict the clean graph, $G^0$, that originated its noisy input, $G^t$. To sample from DiGress, we perform iteratively $T$ reverse steps starting from a fully noisy graph, $G^T$. In each reverse step, the noisy graph is passed as input to the GNN, whose output is then noised back using the posterior term of the diffusion model. Figure adapted from \cite{vignac2022digress}.}
    \label{fig:adapted_digress}
\end{figure}

\paragraph{Diffusion based cell-graph generative model.} 
In our setting, we extend DiGress~\cite{vignac2022digress}, a D3PM adapted for graphs whose nodes and edges have categorical features, to the TLS cell-graph generation setting. This method is a diffusion model composed of two different processes: \emph{forward} and \emph{reverse}. 

In the \emph{forward} process of the diffusion model, we iteratively alter clean graphs, $G^0$, with noise, until we arrive to its fully corrupted form, $G^T$, obtaining a trajectory $\{G^0, \ldots, G^T\}$. The noise model adopted by DiGress is independently employed to nodes and edges through transition matrices. Thus, for each node and for each edge, we apply:
\begin{equation}
    [\mathbf{Q}_X^t]_{ij} = q(x^t=j | x^{t-1}=i) \quad \text{and} \quad [\mathbf{Q}_E^t]_{ij} = q(e^t=j | e^{t-1}=i),
\end{equation}
respectively. Consequently, the categorical distribution that allows us to go one step forward towards the fully noisy graph side of the trajectory is given by: 
\begin{equation}
 q(G^t| G^{t-1}) = (\mathbf{X}^{t-1}\mathbf{Q}_X^{t}, \textbf{E}^{t-1} \mathbf{Q}_E^{t}).
\end{equation}
The adopted noise model follows the marginal scheme~\cite{vignac2022digress}, which promotes transitions to node and edge types that are more prevalent in the training set. Thus, the transition matrices are given by:
\begin{equation}
    \mathbf{Q}_X^t = \alpha^t \mathbf{I} + \beta^t \mathbf{1_A} \mathbf{m}_X' \quad \text{and} \quad \mathbf{Q}_E^t = \alpha^t \mathbf{I} + \beta^t \mathbf{1_B} \mathbf{m}_E',
\end{equation}
where $\alpha^t$ transitions from 1 to 0 with $t$ according to the popular cosine scheduling. Then, $\beta^t = 1-\alpha^t$, $\mathbf{1}_A \in \left\{1\right\}^b, \mathbf{1}_B \in \left\{1\right\}^c$ are filled with ones, and $\mathbf{m}_X' \in \mathbb{R}^b, \mathbf{m}_E' \in \mathbb{R}^c$ are row vectors ($'$ denotes transposition) filled with the marginal distributions of node and edge types, respectively.

In the \emph{reverse} process of DiGress, a graph transformer network (i.e., a type of GNN) is trained to recursively backtrack the trajectories generated by the forward process. To overcome the limited representational power of GNNs, graph-theoretic (cycles and spectral) auxiliary features are added to the transformer model for each reverse step. 

With this setup, DiGress is trained by repeatedly performing the following steps: pick a clean graph from the training set, $G^0$, and perform $t$ forward steps, where $t$ is chosen randomly between $1$ and $T$, obtaining $G^t$. Importantly, it is possible to perform the $t$ forward steps in a single closed-form computation, given $\mathbf{Q}_X^1, \ldots, \mathbf{Q}_X^t, \mathbf{Q}_E^1, \ldots, \mathbf{Q}_E^t$. Then, the GNN, which is the only trainable component of DiGress, is fed with $G^t$ and predicts probability distributions over the nodes and edges types for each node and edge of $G^0$, being then subject to binary cross-entropy loss. After having trained the GNN, we can sample from DiGress. First, we sample a fully noisy graph, $G^T$, from the limit distribution of the forward process, which is fixed by definition. Then, to obtain a new clean sample, we iteratively denoise $G^T$ by performing $T$ reverse steps. Each of the reverse steps consists of a GNN prediction given the noisy graph as input, followed by a "noise back" step using the posterior term of DiGress. This posterior term can be computed in closed-form provided the diffusion model transition matrices. Both the training and the sampling procedures from DiGress are illustrated in \Cref{fig:adapted_digress}.

Altogether, the diffusion-based graph generative model built on Digress permits to infer the distribution of the data that the model has been trained on. As a result, when properly trained on actual samples of TLS cell-graphs, the generation of novel graphs is simply achieved by sampling from the learned cell-graph distribution.

%% file: content/experiments.tex
\section{Experiments}\label{sec:experiments}

In this section, we first describe the data preprocessing steps, followed by an evaluation of the quality of the cell-graphs generated using DiGress, comparing it to a non deep learning baseline. Then, we illustrate the utility of the trained DiGress models for data augmentation in a TLS content classification task. Finally, we report the computational resources required for the previous tasks.

\paragraph{Data and preprocessing.}

We create our actual datasets\footnote{The used datasets resulted from a collaboration with Centre Hospitalier Universitaire Vaudois and are not publicly available.} following the classical procedure for cell-graph extraction in digital pathology~\cite{gunduz2004cell,schaadt2020graph}. The first step consists of segmenting the cells of the WSI. We set the detected cells to nodes, whose types form node features. We consider 9 different cell types, identified in \Cref{fig:cell_graph}. This imposes $b=9$ in the generative model. To obtain the whole slide graph, edges are set between adjacent nodes using Delaunay triangulation and edge thresholding (edges longer than 30 $\mu$m are ignored). As we only consider one edge type, we have $c=1$. Finally, from each whole slide graph, we extract several 4-hop subgraphs centered on B-cells. These subgraphs are then assigned to two datasets: $\mathcal{D}^{\sim\textrm{TLS}}_\textrm{Real}$ composed of cell-graphs with low TLS content ($\ka{1}<0.05$) and $\mathcal{D}^{\textrm{TLS}}_\textrm{Real}$ with high TLS content cell-graphs ($\ka{2}>0.05$). $\mathcal{D}^{\sim\textrm{TLS}}_\textrm{Real}$ and $\mathcal{D}^{\textrm{TLS}}_\textrm{Real}$ contain 3020 and 5042 cell-graphs, respectively, with varying number of nodes between 20 and 103. We also define $\mathcal{D}_\textrm{Real} =\mathcal{D}^{\sim\textrm{TLS}}_\textrm{Real} \cup \mathcal{D}^{\textrm{TLS}}_\textrm{Real}$.

\paragraph{Baseline.} We define a non deep learning based but reasonable baseline that captures the 1-hop dependencies of cell-graphs. This model learns the following marginal distributions from the respective training set: number of nodes per graph (distribution \emph{I}), cell types (distribution \emph{II}), and edge types given the cell types of the edge vertices (distribution \emph{III}). Importantly, in the scope of this paper, \emph{I} and \emph{II} are categorical distributions, while \emph{III} consists of a Bernoulli distribution. Thereafter, the sampling procedure of a graph from the trained baseline model is as follows: \emph{i)} sample a number of nodes from \emph{I}; \emph{ii)} for each of those nodes, sample a cell type from \emph{II}; and, finally, \emph{iii)} sample an edge type between every two nodes from \emph{III}. Noteworthy, for a given edge, the vertices cell types are, at that point, already known from step \emph{ii}).

\paragraph{Evaluation of synthetic cell-graphs distributions.}
We separately train a DiGress model and a baseline model for each of the datasets $\mathcal{D}^{\sim\textrm{TLS}}_\textrm{Real}$ and $\mathcal{D}^{\textrm{TLS}}_\textrm{Real}$, yielding a total of 4 trained models.
We sample from each of the trained models, obtaining $\mathcal{D}^{\sim\textrm{TLS}}_\textrm{DiGress}$, $\mathcal{D}^{\textrm{TLS}}_\textrm{DiGress}$, $\mathcal{D}^{\sim\textrm{TLS}}_\textrm{Baseline}$, and $\mathcal{D}^{\textrm{TLS}}_\textrm{Baseline}$, each with 5000 cell-graphs. Then, we compare these generated datasets with the respective training sets through their distributions, in a biologically meaningful way: using the TLS embedding distributions. This comparison is made through four different metrics typically adopted in synthetic data evaluation~\cite{qian2023synthcity}, as they emphasize different similarities between probability distributions. In particular, the Kolmogorov-Smirnov test (KS) focuses on similarity of distributions based on the maximum vertical distance between their cumulative distribution functions. The Wasserstein distance (WD) captures differences in the shape, location, and spread of distributions by measuring the minimum amount of work required for transformation of one into the other. The Jensen-Shannon divergence (D$_\textrm{JS}$) captures dissimilarity based on information content by comparing the divergence of each distribution from their average distribution. Lastly, the maximum mean discrepancy (MMD) evaluates the difference in means of data representations transformed using a kernel function. The results, presented in \Cref{tab:dist_metrics}, show that the intricate biological dependencies found in a cell-graph go beyond 1-hop relations, since those are the ones that the baseline is able to capture accurately. In contrast, by capturing higher order and more complex dependencies, DiGress significantly outperforms the baseline in both training sets. This fact highlights the pertinence of graph-based diffusion models to capture cell-graph distributions.

\begin{table}[t!]
\centering
\caption{Evaluation of the generated datasets through TLS embedding distribution comparison with the real TLS dataset, $\mathcal{D}^{\textrm{TLS}}_\textrm{Real}$, (\emph{left}) and $\sim$TLS dataset, $\mathcal{D}^{\sim\textrm{TLS}}_\textrm{Real}$, (\emph{right}). The TLS embedding entries are denoted by $\ka{a}$. KS stands for Kolmogorov-Smirnov test, WD for Wasserstein distance, D$_\textrm{JS}$ for Jensen-Shannon divergence, and MMD for maximum mean discrepancy.}
\label{tab:dist_metrics}
\resizebox{\textwidth}{!}{%

\begin{tabular}{@{}ll|rrrrrr|rrrrrr@{}}
\toprule
 &
   &
  \multicolumn{6}{c|}{TLS} &
  \multicolumn{6}{c}{$\sim$TLS} \\
 &
   &
  \multicolumn{1}{c}{$\ka{0}$} &
  \multicolumn{1}{c}{$\ka{1}$} &
  \multicolumn{1}{c}{$\ka{2}$} &
  \multicolumn{1}{c}{$\ka{3}$} &
  \multicolumn{1}{c}{$\ka{4}$} &
  \multicolumn{1}{c|}{$\ka{5}$} &
  \multicolumn{1}{c}{$\ka{0}$} &
  \multicolumn{1}{c}{$\ka{1}$} &
  \multicolumn{1}{c}{$\ka{2}$} &
  \multicolumn{1}{c}{$\ka{3}$} &
  \multicolumn{1}{c}{$\ka{4}$} &
  \multicolumn{1}{c}{$\ka{5}$} \\ \midrule
\multirow{2}{*}{KS} &
  Baseline &
  0.127 &
  0.092 &
  \textbf{0.088} &
  0.105 &
  0.114 &
  0.080 &
  0.136 &
  0.398 &
  0.104 &
  0.018 &
  0.002 &
  \textbf{0.000} \\
 &
  DiGress &
  \textbf{0.060} &
  \textbf{0.062} &
  0.110 &
  \textbf{0.051} &
  \textbf{0.049} &
  \textbf{0.057} &
  \textbf{0.097} &
  \textbf{0.160} &
  \textbf{0.007} &
  \textbf{0.000} &
  \textbf{0.000} &
  \textbf{0.000} \\ \midrule
\multirow{2}{*}{WD} &
  Baseline &
  0.028 &
  \textbf{0.041} &
  \textbf{0.044} &
  0.034 &
  0.016 &
  0.009 &
  0.064 &
  0.079 &
  0.013 &
  0.002 &
  0.000 &
  \textbf{0.000} \\
 &
  DiGress &
  \textbf{0.019} &
  0.044 &
  0.045 &
  \textbf{0.024} &
  \textbf{0.016} &
  \textbf{0.006} &
  \textbf{0.053} &
  \textbf{0.017} &
  \textbf{0.001} &
  \textbf{0.000} &
  \textbf{0.000} &
  \textbf{0.000} \\ \midrule
\multirow{2}{*}{D$_\textrm{JS}$} &
  Baseline &
  0.199 &
  0.202 &
  0.228 &
  0.144 &
  0.120 &
  0.084 &
  0.168 &
  0.419 &
  0.190 &
  0.079 &
  0.028 &
  \textbf{0.000} \\
 &
  DiGress &
  \textbf{0.087} &
  \textbf{0.126} &
  \textbf{0.215} &
  \textbf{0.093} &
  \textbf{0.088} &
  \textbf{0.071} &
  \textbf{0.136} &
  \textbf{0.265} &
  \textbf{0.053} &
  \textbf{0.000} &
  \textbf{0.000} &
  \textbf{0.000} \\ \midrule
\multirow{2}{*}{MMD} &
  Baseline &
  0.062 &
  0.076 &
  0.071 &
  0.068 &
  0.049 &
  0.024 &
  0.095 &
  0.602 &
  0.059 &
  0.002 &
  0.000 &
  \textbf{0.000} \\
 &
  DiGress &
  \textbf{0.012} &
  \textbf{0.024} &
  \textbf{0.033} &
  \textbf{0.013} &
  \textbf{0.012} &
  \textbf{0.013} &
  \textbf{0.051} &
  \textbf{0.086} &
  \textbf{0.000} &
  \textbf{0.000} &
  \textbf{0.000} &
  \textbf{0.000} \\ \bottomrule
\end{tabular}%
}
\end{table}

\paragraph{Synthetic cell-graphs in data augmentation.}

To illustrate the utility of generative models in downstream tasks, we explore data augmentation with synthetic cell-graphs in a scarcity regime of real data, a setting often found in biomedical applications. 
In particular, we consider a binary classification task where a graph convolutional network (GCN) is trained to predict if a cell-graph has a high or low TLS content. We build two datasets with freshly collected cell-graphs from the same tissue slide cell-graphs, $\mathcal{D}'^{\sim\textrm{TLS}}_\textrm{Real}$ and $\mathcal{D}'^{\textrm{TLS}}_\textrm{Real}$, both with 100 cell-graphs.
We split the dataset $\mathcal{D}'_\textrm{Real} =\mathcal{D}'^{\sim\textrm{TLS}}_\textrm{Real} \cup \mathcal{D}'^{\textrm{TLS}}_\textrm{Real}$ into balanced and equally sized training and test sets. Finally, during the training of the GCN, we augment the training set with positive and negative cell-graphs (by the same amount) coming from three different origins: \emph{i)} real data, $\mathcal{D}^{\sim\textrm{TLS}}_\textrm{Real}$ and $\mathcal{D}^{\textrm{TLS}}_\textrm{Real}$; \emph{ii)} synthetic cell-graphs generated by DiGress, $\mathcal{D}^{\sim\textrm{TLS}}_\textrm{DiGress}$, and $\mathcal{D}^{\textrm{TLS}}_\textrm{DiGress}$; and \emph{iii)} 
synthetic cell-graphs generated by the baseline model, $\mathcal{D}^{\sim\textrm{TLS}}_\textrm{Baseline}$, and $\mathcal{D}^{\textrm{TLS}}_\textrm{Baseline}$. The magnitude of data augmentation ranges from $1\times$ (which means no augmentation) to $40\times$ (which means that the augmented training set is 40 times larger than the original training set).

Since the adopted criteria to consider high and low TLS content cell-graphs consists of a 2-hop condition, we consider a 2-layer GCN for the illustrative binary graph classification task. This model is trained using binary cross-entropy until a maximum of 5000 epochs, or less in case the cross-entropy of the model does not increase in the validation set for 100 epochs (early-stopping). For each type and magnitude of augmentation, we tune the hyperparameters within a grid: learning rate takes the values 0.01, 0.001, or 0.0001; the dimension of the graph embeddings in the hidden layers is 8 or 16; the dropout rate is 0.2 or 0.5. We find the best hyperparameter combination considering the mean AUROC of the trained models in the corresponding validation sets in a 5-fold stratified cross-validation procedure. The 5 models that share the best hyperparameter configuration are then evaluated at the test set. The obtained results can be found in \Cref{tab:data_aug}. 

\begin{table}[t!]
\centering
\caption{Validation (top) and test (bottom) AUROC (mean ± standard error of the mean) obtained for the 5 models sharing the best hyperparameter combination for each type and magnitude of data augmentation.}
\label{tab:data_aug}
\resizebox{\textwidth}{!}{%
\begin{tabular}{@{}lrrrrrrr@{}}
\toprule
\textbf{Val.} &
  \multicolumn{1}{c}{1$\times$} &
  \multicolumn{1}{c}{2$\times$} &
  \multicolumn{1}{c}{3$\times$} &
  \multicolumn{1}{c}{5$\times$} &
  \multicolumn{1}{c}{10$\times$} &
  \multicolumn{1}{c}{20$\times$} &
  \multicolumn{1}{c}{40$\times$} \\ \midrule
Real &
  \textbf{0.970} \tiny{± 0.010} &
  \textbf{0.976} \tiny{± 0.008} &
  \textbf{0.982} \tiny{± 0.009} &
  \textbf{0.990} \tiny{± 0.006} &
  \textbf{0.990} \tiny{± 0.005} &
  \textbf{0.996} \tiny{± 0.004} &
  \textbf{0.994} \tiny{± 0.004} \\
DiGress &
  \textbf{0.970} \tiny{± 0.010} &
  0.972  \tiny{± 0.015} &
  0.978  \tiny{± 0.013} &
  0.982  \tiny{± 0.011} &
  0.974  \tiny{± 0.011} &
  0.976  \tiny{± 0.017} &
  0.978  \tiny{± 0.015} \\
Baseline &
  \textbf{0.970} \tiny{± 0.010} &
  0.960  \tiny{± 0.019} &
  0.944  \tiny{± 0.028} &
  0.950  \tiny{± 0.024} &
  0.932  \tiny{± 0.032} &
  0.934  \tiny{± 0.031} &
  0.938  \tiny{± 0.023} \\ \bottomrule
\end{tabular}%
}
\resizebox{\textwidth}{!}{%

\begin{tabular}{@{}lrrrrrrr@{}}
\toprule
\textbf{Test} &
  \multicolumn{1}{c}{1$\times$} &
  \multicolumn{1}{c}{2$\times$} &
  \multicolumn{1}{c}{3$\times$} &
  \multicolumn{1}{c}{5$\times$} &
  \multicolumn{1}{c}{10$\times$} &
  \multicolumn{1}{c}{20$\times$} &
  \multicolumn{1}{c}{40$\times$} \\ \midrule
Real &
  \textbf{0.921} \tiny{± 0.004} &
  0.926 \tiny{± 0.005} &
  \textbf{0.952} \tiny{± 0.006} &
  \textbf{0.947} \tiny{± 0.003} &
  \textbf{0.956} \tiny{± 0.010} &
  \textbf{0.962} \tiny{± 0.008} &
  \textbf{0.966} \tiny{± 0.010} \\
DiGress &
  \textbf{0.921} \tiny{± 0.004} &
  \textbf{0.946} \tiny{± 0.007} &
  0.943 \tiny{± 0.010} &
  0.947 \tiny{± 0.010} &
  0.945 \tiny{± 0.010} &
  0.931 \tiny{± 0.009} &
  0.939 \tiny{± 0.009} \\
Baseline &
  \textbf{0.921} \tiny{± 0.004} &
  0.938 \tiny{± 0.009} &
  0.927 \tiny{± 0.007} &
  0.932 \tiny{± 0.005} &
  0.928 \tiny{± 0.008} &
  0.927 \tiny{± 0.007} &
  0.922 \tiny{± 0.004} \\ \bottomrule
\end{tabular}%
}
\end{table}

As a general observation, data augmentation leads to improvements in model performance, which is a consequence of the regularization effect of data augmentation in learning tasks. 
However, this improvement differs with the origin of the added cell-graphs. 
For real data, we observe that overall the stronger the augmentation magnitude, the better the performance. This result is an expected consequence of the augmentation cell-graphs following exactly the same distribution as the ones found in the training and test set: despite being different, all of them were extracted from the same tissue slides. 
When the augmentation is performed with DiGress generated cell-graphs, the model test performance increases as well, but not as much as by using real data. 
This result comes as a consequence of the samples generated by DiGress not following exactly the same distribution of $\mathcal{D}_\textrm{Real}$ (or, equivalently, $\mathcal{D}'_\textrm{Real}$), as shown in \Cref{tab:dist_metrics}. 
By the same token, the performance improvement brought by the augmentation with the baseline model cell-graphs is more subtle. 
We can finally note that, for the augmentations with synthetically generated cell-graphs, there is no monotonic trend of performance improvement with the magnitude of data augmentation. We attribute the test AUROC deterioration for augmentations larger than $5\times$ for DiGress and $2\times$ for the baseline model to an excessive exposition of the model to synthetic data, which unavoidably leads to distribution shifts in the total training set. The performance deterioration is detected at smaller augmentation magnitudes for the baseline model as it causes larger distribution shifts.

\paragraph{Computational resources.}  
The training of the deep learning models was carried out using a single Tesla V100 GPU. The training of the DiGress models on the $\mathcal{D}^{\sim\textrm{TLS}}_\textrm{Real}$ and $\mathcal{D}^{\textrm{TLS}}_\textrm{Real}$ datasets required 19 and 30 hours, respectively. The generation of $\mathcal{D}^{\sim\textrm{TLS}}_\textrm{DiGress}$ and $\mathcal{D}^{\textrm{TLS}}_\textrm{DiGress}$ consumed 17 hours each, approximately. This sampling procedure was performed in the same single GPU. For the baseline model, its training on the $\mathcal{D}^{\sim\textrm{TLS}}_\textrm{Real}$ and $\mathcal{D}^{\textrm{TLS}}_\textrm{Real}$ datasets only took 14 and 20 minutes, respectively. The generation of $\mathcal{D}^{\sim\textrm{TLS}}_\textrm{Baseline}$ and $\mathcal{D}^{\textrm{TLS}}_\textrm{Baseline}$ requires only 20 seconds for each of them. Finally, for the training of the GCN used in the classification task, 1260 runs were carried out to cover all the hyperparameter configurations explored, cross-validation folds, and sources of augmentation. The longest of them took approximately 17 minutes.

%% file: content/conclusion.tex
\section{Conclusion}

This paper provides, to the best of our knowledge, the first application of graph-based diffusion models to the digital pathology domain. We leverage on a current state-of-the-art diffusion model for graphs with categorical node and edges features, DiGress, to successfully generate high and low TLS content cell-graphs. Our results highlight the relevance of capturing higher order relations to accurately model the complex biological dependencies that rule the digital pathology data. Furthermore, we demonstrate the practical value of generative models as a means of augmenting this data type.
As future work, we are looking at properly incorporating \emph{prior} biological knowledge to the generative model in order to further enhance the promising results obtained in this work. The main limitation of the application of current state-of-the-art graph-based generative models in whole-slide pathology images is related with their limited scalability: developing implementations that can generate large graphs is an important future direction.

%% file: content/acknowledgements.tex
\section{Acknowledgements}

The authors would like to thank Alexandre Wicky, Michel A. Cuendet and Olivier Michielin for the fruitful discussions about the biological interpretation of cell-graph representations and to Beril Besbinar for the help in data pre-processing and useful methodological suggestions.  

%% file: content/appendix.tex
\appendix

\section{Examples of high and low TLS content subgraph}

\begin{figure}[h]
    \includegraphics[width=0.45\textwidth]{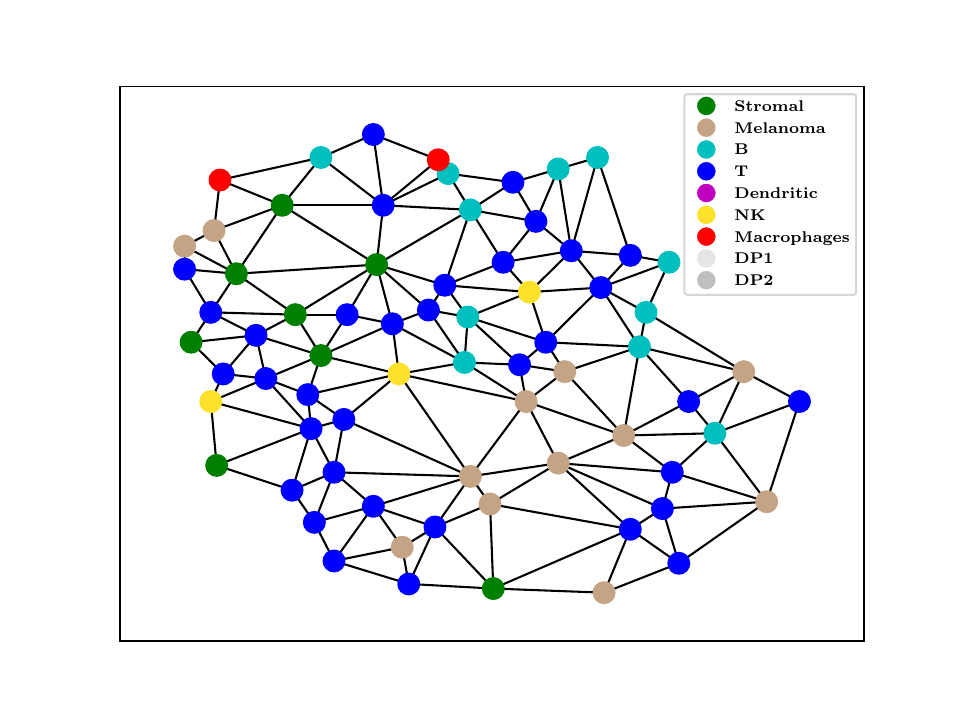}
    \includegraphics[width=0.45\textwidth]{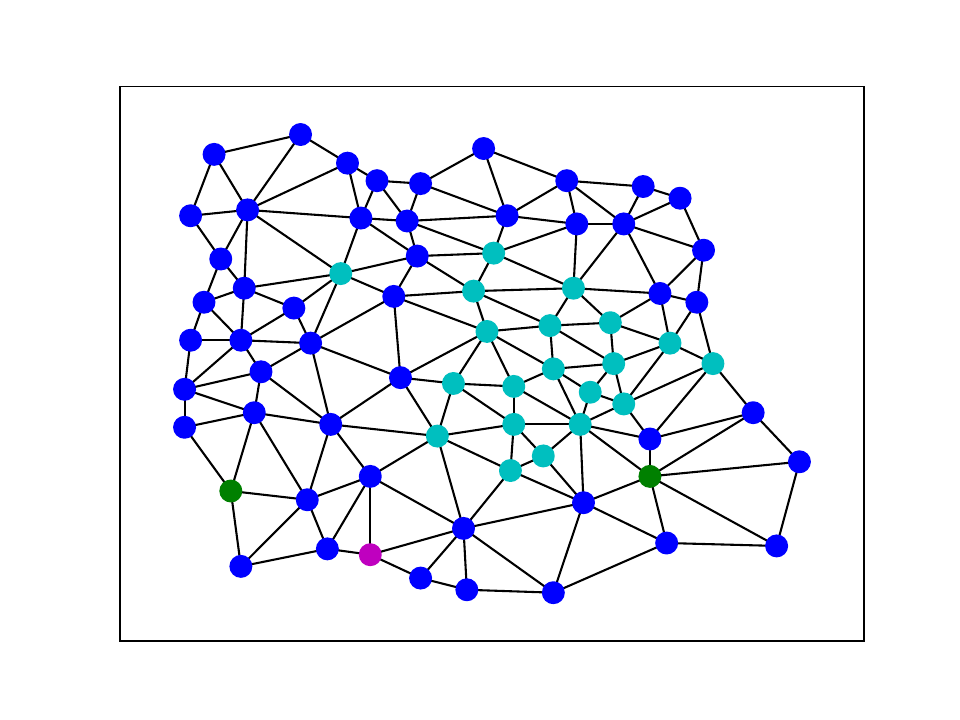}
    \includegraphics[width=0.45\textwidth]{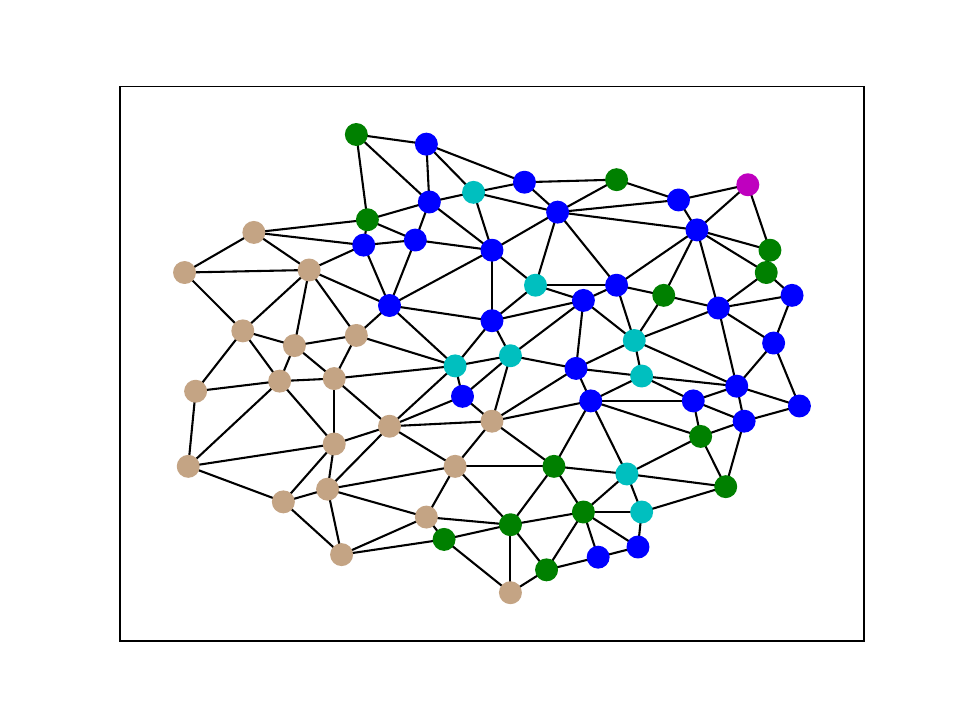}
    \includegraphics[width=0.45\textwidth]{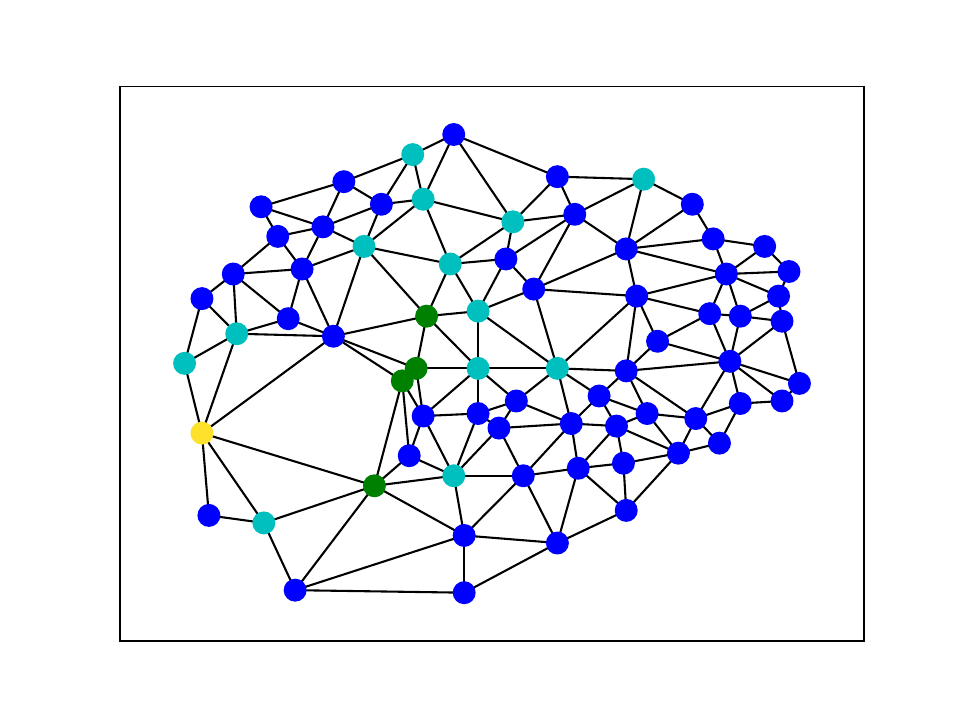}
    \includegraphics[width=0.45\textwidth]{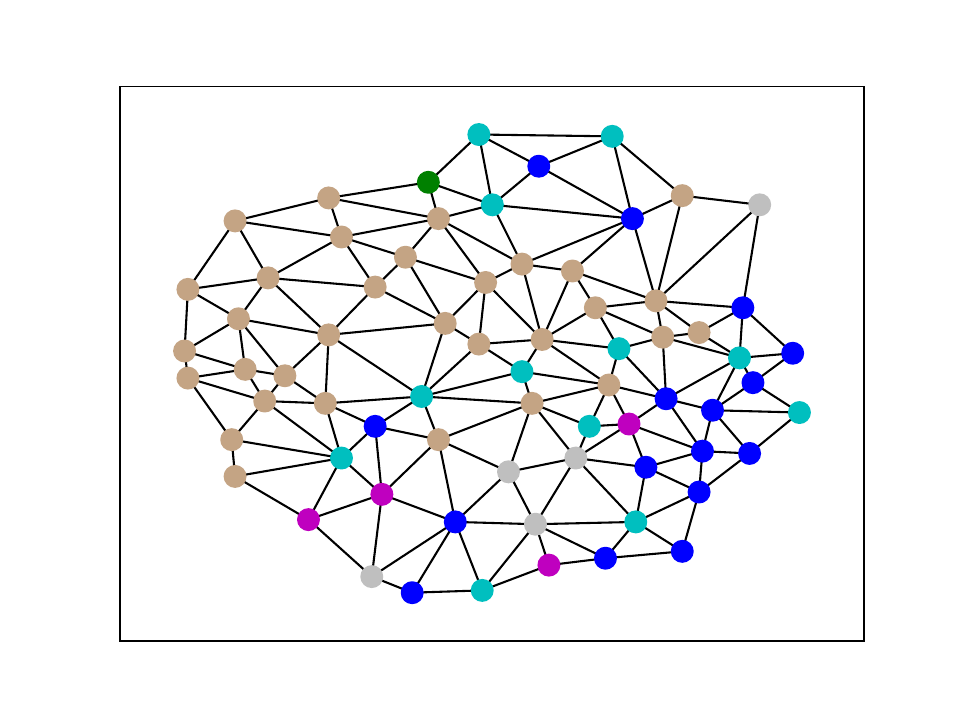}
    \includegraphics[width=0.45\textwidth]{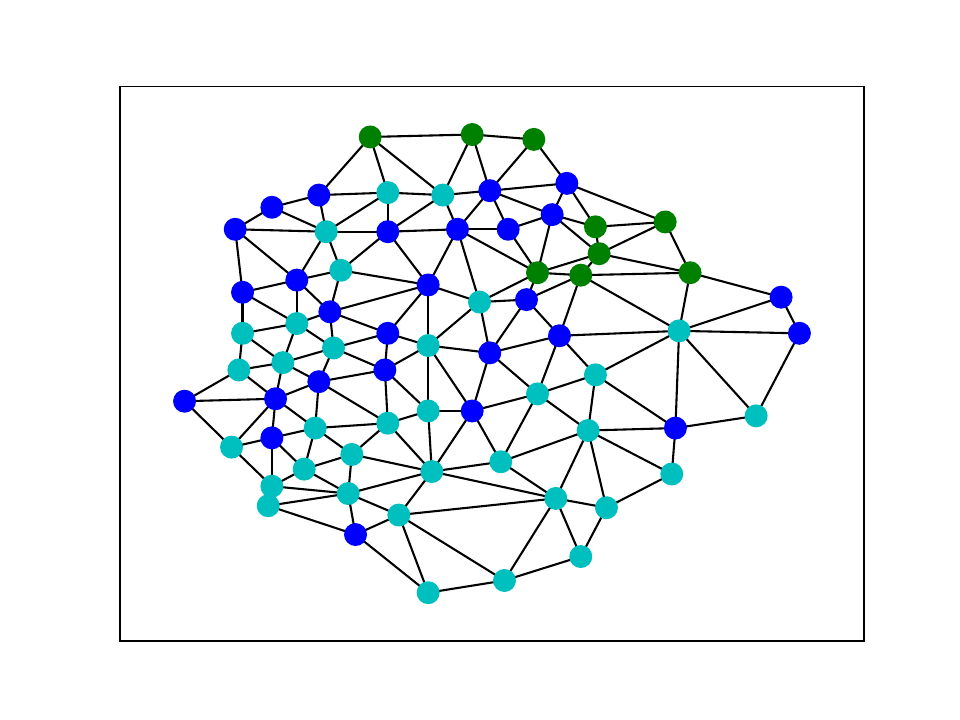}
    \centering
    \caption{Subgraphs extracted from whole-slide cell-graphs.
    \emph{Left:} Low TLS content cell-graphs.
    \emph{Right:} High TLS content cell-graphs. These cell-graphs are similar to the one found at the center of \Cref{fig:cell_graph}.
    }
    \label{fig:high_vs_low_tls_graphs}
\end{figure}

In this section we present examples of high and low TLS content cell-graphs to provide further intuition on the difference between the two classes. It is possible to observe a distinct cluster of B-cells surrounded by supporting T-cells for high TLS content cell-graphs, whereas the low TLS content counterparts have a more heterogeneous composition and configuration, as illustrated in \Cref{fig:high_vs_low_tls_graphs}.

\section{Distributions of TLS embeddings}\label{app:dist_embeddings}

\begin{figure}[h]
    \includegraphics[width=0.45\textwidth]{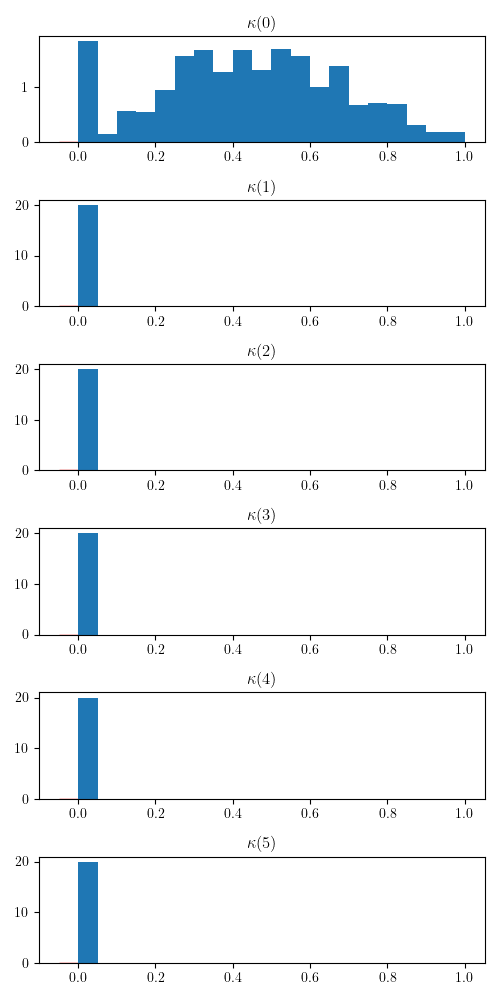}
    \includegraphics[width=0.45\textwidth]{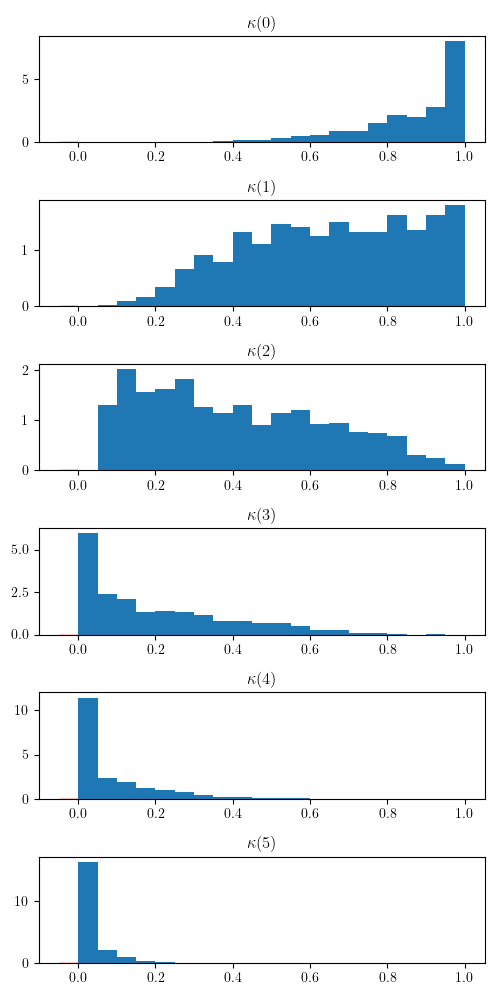}
    \centering
    \caption{Empirical distributions obtained for the TLS embeddings in the real data.
    \emph{Left:} TLS embedding distributions for $\mathcal{D}^{\sim\textrm{TLS}}_\textrm{Real}$.
    \emph{Right:} TLS embedding distributions for $\mathcal{D}^{\textrm{TLS}}_\textrm{Real}$.
    }
    \label{fig:dist_emb_groundtruth}
\end{figure}

\begin{figure}[t!]
    \includegraphics[width=0.45\textwidth]{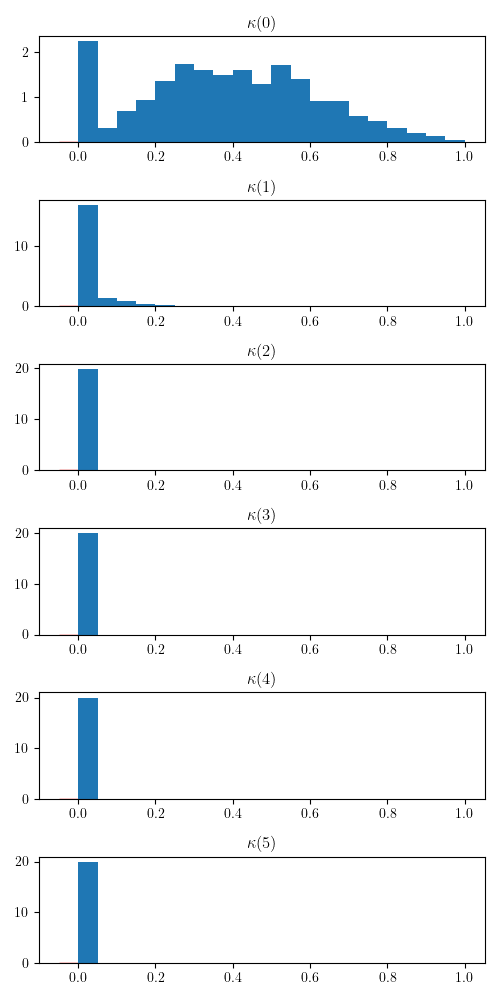}
    \includegraphics[width=0.45\textwidth]{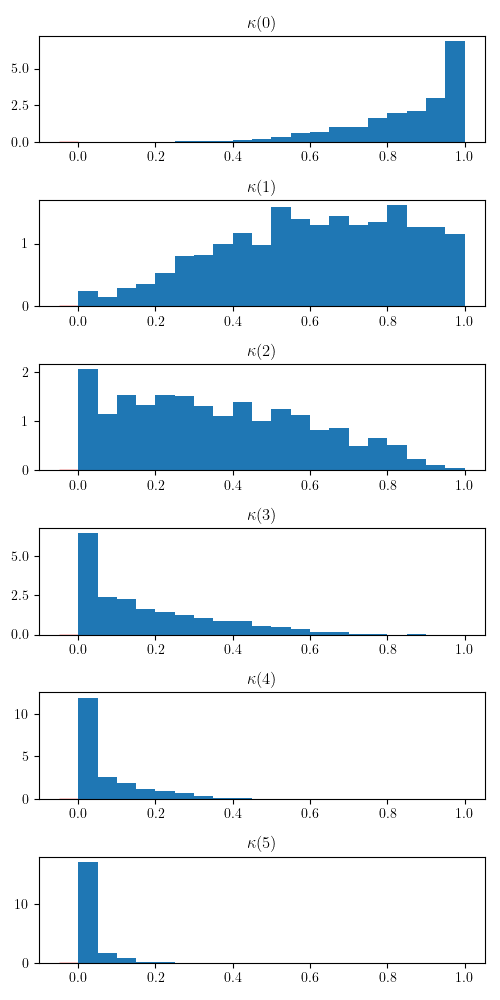}
    \centering
    \caption{Empirical distributions obtained for the TLS embeddings in the dataset generated with DiGress.
    \emph{Left:} TLS embedding distributions for $\mathcal{D}^{\sim\textrm{TLS}}_\textrm{DiGress}$.
    \emph{Right:} TLS embedding distributions for $\mathcal{D}^{\textrm{TLS}}_\textrm{DiGress}$.
    }
    \label{fig:dist_emb_digress}
\end{figure}

\begin{figure}[t!]
    \includegraphics[width=0.45\textwidth]{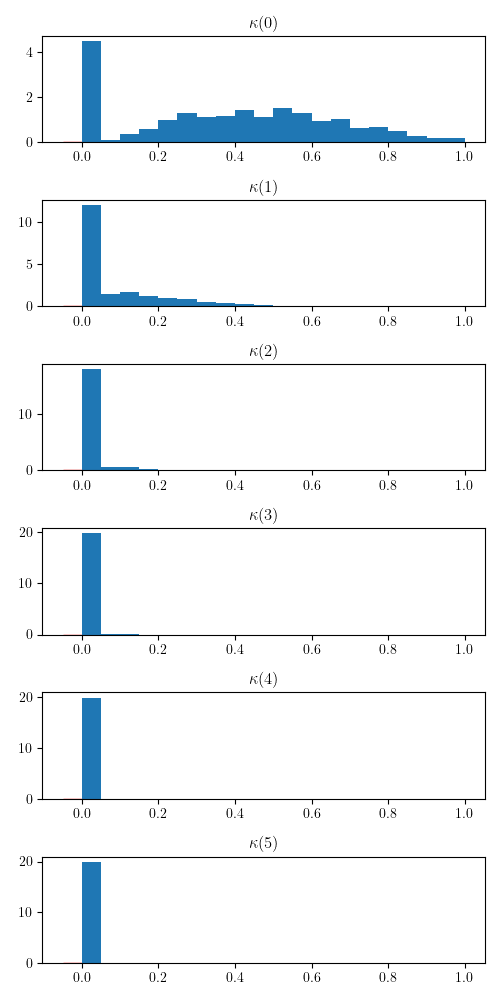}
    \includegraphics[width=0.45\textwidth]{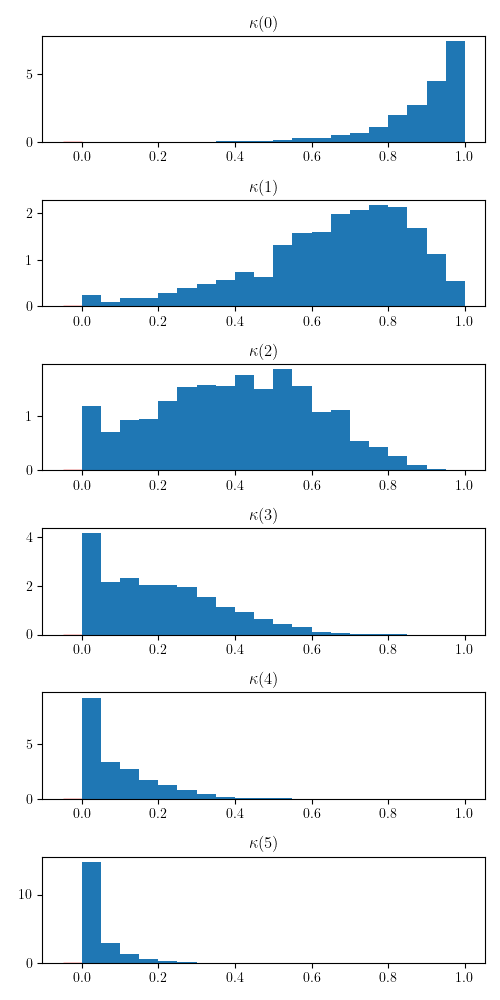}
    \centering
    \caption{Empirical distributions obtained for the TLS embeddings in the dataset generated with the baseline model.
    \emph{Left:} TLS embedding distributions for $\mathcal{D}^{\sim\textrm{TLS}}_\textrm{Baseline}$.
    \emph{Right:} TLS embedding distributions for $\mathcal{D}^{\textrm{TLS}}_\textrm{Baseline}$.
    }
    \label{fig:dist_emb_naive}
\end{figure}

To further illustrate the results obtained in \Cref{sec:experiments}, we provide the TLS embeddings distributions for the different datasets considered. The distributions obtained for the large real dataset, for the dataset sampled from DiGress, and for the dataset sampled from the baseline model are depicted in \Cref{fig:dist_emb_groundtruth}, \Cref{fig:dist_emb_digress}, and \Cref{fig:dist_emb_naive}, respectively. 

In \Cref{fig:dist_emb_groundtruth}, it is possible to observe the sharp transitions imposed by the low TLS content criterion ($\ka{1}<0.05$) for the $\mathcal{D}^{\sim\textrm{TLS}}_\textrm{Real}$ dataset and by the high TLS content criterion ($\ka{2}>0.05$) for the $\mathcal{D}^{\textrm{TLS}}_\textrm{Real}$. This transition is not well captured by the generative models, explaining their worse performance for $\ka{1}$ (in $\sim$TLS) and $\ka{2}$ (in TLS), namely for DiGress (see \Cref{tab:dist_metrics}).

Furthermore, it is possible to observe that the distributions tend to collapse to 0 with the increase of the index, $a$, of the TLS embedding, $\ka{a}$. In fact, by definition, for a given graph, the TLS embedding entries are decreasing with its index (see \Cref{eq:tls_embedding}). Therefore, it is possible to conclude from \Cref{fig:dist_emb_groundtruth} that, in the vast majority of the cases, no relevant information remains in $\ka{a}$ for $a>5$, supporting our decision of discarding their computation.

%% file: paper.bbl
\begin{thebibliography}{10}
\providecommand{\url}[1]{\texttt{#1}}
\providecommand{\urlprefix}{URL }
\providecommand{\doi}[1]{https://doi.org/#1}

\bibitem{ahmedt2022survey}
Ahmedt-Aristizabal, D., Armin, M.A., Denman, S., Fookes, C., Petersson, L.: {A survey on graph-based deep learning for computational histopathology}. Computerized Medical Imaging and Graphics  \textbf{95},  102027 (2022)

\bibitem{austin2021structured}
Austin, J., Johnson, D.D., Ho, J., Tarlow, D., van~den Berg, R.: {Structured denoising diffusion models in discrete state-spaces}. In: {{Advances in Neural Information Processing Systems}} (2021)

\bibitem{bera2019artificial}
Bera, K., Schalper, K.A., Rimm, D.L., Velcheti, V., Madabhushi, A.: {Artificial intelligence in digital pathology—new tools for diagnosis and precision oncology}. Nature reviews Clinical oncology  \textbf{16}(11),  703--715 (2019)

\bibitem{chlap2021review}
Chlap, P., Min, H., Vandenberg, N., Dowling, J., Holloway, L., Haworth, A.: {A review of medical image data augmentation techniques for deep learning applications}. Journal of Medical Imaging and Radiation Oncology  \textbf{65}(5),  545--563 (2021)

\bibitem{de2002modeling}
De~Jong, H.: {Modeling and simulation of genetic regulatory systems: a literature review}. Journal of computational biology  \textbf{9}(1),  67--103 (2002)

\bibitem{dieu2016tertiary}
Dieu-Nosjean, M.C., Giraldo, N.A., Kaplon, H., Germain, C., Fridman, W.H., Saut{\`e}s-Fridman, C.: {Tertiary lymphoid structures, drivers of the anti-tumor responses in human cancers}. Immunological reviews  \textbf{271}(1),  260--275 (2016)

\bibitem{gunduz2004cell}
Gunduz, C., Yener, B., Gultekin, S.H.: {The cell graphs of cancer}. Bioinformatics  \textbf{20}(suppl\_1),  i145--i151 (2004)

\bibitem{haefeli2022diffusion}
Haefeli, K.K., Martinkus, K., Perraudin, N., Wattenhofer, R.: {Diffusion Models for Graphs Benefit From Discrete State Spaces}. In: NeurIPS 2022 Workshop: New Frontiers in Graph Learning (2022)

\bibitem{van2002graph}
van Helden, J., Wernisch, L., Gilbert, D., Wodak, S.: {Graph-based analysis of metabolic networks}. Bioinformatics and genome analysis pp. 245--274 (2002)

\bibitem{helmink2020b}
Helmink, B.A., Reddy, S.M., Gao, J., Zhang, S., Basar, R., Thakur, R., Yizhak, K., Sade-Feldman, M., Blando, J., Han, G., et~al.: {B cells and tertiary lymphoid structures promote immunotherapy response}. Nature  \textbf{577}(7791),  549--555 (2020)

\bibitem{ho2020denoising}
Ho, J., Jain, A., Abbeel, P.: {Denoising diffusion probabilistic models}. In: {{Advances in Neural Information Processing Systems}} (2020)

\bibitem{jaume2021histocartography}
Jaume, G., Pati, P., Anklin, V., Foncubierta, A., Gabrani, M.: {HistoCartography: A toolkit for graph analytics in digital pathology}. In: MICCAI Workshop on Computational Pathology. PMLR (2021)

\bibitem{jaume2021quantifying}
Jaume, G., Pati, P., Bozorgtabar, B., Foncubierta, A., Anniciello, A.M., Feroce, F., Rau, T., Thiran, J.P., Gabrani, M., Goksel, O.: {Quantifying explainers of graph neural networks in computational pathology}. In: {{IEEE Conference on Computer Vision and Pattern Recognition}} (2021)

\bibitem{jo2022score}
Jo, J., Lee, S., Hwang, S.J.: {Score-based generative modeling of graphs via the system of stochastic differential equations}. In: {{International Conference on Machine Learning}}. PMLR (2022)

\bibitem{jose2021generative}
Jose, L., Liu, S., Russo, C., Nadort, A., Di~Ieva, A.: {Generative adversarial networks in digital pathology and histopathological image processing: A review}. Journal of Pathology Informatics  \textbf{12}(1), ~43 (2021)

\bibitem{lee2016tertiary}
Lee, H.J., Park, I.A., Song, I.H., Shin, S.J., Kim, J.Y., Yu, J.H., Gong, G.: {Tertiary lymphoid structures: prognostic significance and relationship with tumour-infiltrating lymphocytes in triple-negative breast cancer}. Journal of clinical pathology  \textbf{69}(5),  422--430 (2016)

\bibitem{li2022graph}
Li, M.M., Huang, K., Zitnik, M.: {Graph representation learning in biomedicine and healthcare}. Nature Biomedical Engineering  \textbf{6}(12),  1353--1369 (2022)

\bibitem{moghadam2023morphology}
Moghadam, P.A., Van~Dalen, S., Martin, K.C., Lennerz, J., Yip, S., Farahani, H., Bashashati, A.: {A morphology focused diffusion probabilistic model for synthesis of histopathology images}. In: Proceedings of the IEEE/CVF Winter Conference on Applications of Computer Vision (2023)

\bibitem{munoz2020tertiary}
Munoz-Erazo, L., Rhodes, J.L., Marion, V.C., Kemp, R.A.: {Tertiary lymphoid structures in cancer--considerations for patient prognosis}. Cellular \& molecular immunology  \textbf{17}(6),  570--575 (2020)

\bibitem{niu2020permutation}
Niu, C., Song, Y., Song, J., Zhao, S., Grover, A., Ermon, S.: {Permutation invariant graph generation via score-based generative modeling}. In: International Conference on Artificial Intelligence and Statistics. PMLR (2020)

\bibitem{pitzalis2014ectopic}
Pitzalis, C., Jones, G.W., Bombardieri, M., Jones, S.A.: {Ectopic lymphoid-like structures in infection, cancer and autoimmunity}. Nature Reviews Immunology  \textbf{14}(7),  447--462 (2014)

\bibitem{qian2023synthcity}
Qian, Z., Cebere, B.C., van~der Schaar, M.: {Synthcity: facilitating innovative use cases of synthetic data in different data modalities}. arXiv preprint arXiv:2301.07573  (2023)

\bibitem{schaadt2020graph}
Schaadt, N.S., Sch{\"o}nmeyer, R., Forestier, G., Brieu, N., Braubach, P., Nekolla, K., Meyer-Hermann, M., Feuerhake, F.: {Graph-based description of tertiary lymphoid organs at single-cell level}. PLoS Computational Biology  \textbf{16}(2),  e1007385 (2020)

\bibitem{serag2019translational}
Serag, A., Ion-Margineanu, A., Qureshi, H., McMillan, R., Saint~Martin, M.J., Diamond, J., O'Reilly, P., Hamilton, P.: {Translational AI and deep learning in diagnostic pathology}. Frontiers in medicine  \textbf{6}, ~185 (2019)

\bibitem{sohl2015deep}
Sohl-Dickstein, J., Weiss, E., Maheswaranathan, N., Ganguli, S.: {Deep unsupervised learning using nonequilibrium thermodynamics}. In: {{International Conference on Machine Learning}}. PMLR (2015)

\bibitem{tang2021augmenting}
Tang, C., Vishwakarma, S., Li, W., Adve, R., Julier, S., Chetty, K.: {Augmenting experimental data with simulations to improve activity classification in healthcare monitoring}. In: 2021 IEEE radar conference (RadarConf21). IEEE (2021)

\bibitem{vignac2022digress}
Vignac, C., Krawczuk, I., Siraudin, A., Wang, B., Cevher, V., Frossard, P.: {DiGress: Discrete Denoising diffusion for graph generation}. In: {{International Conference on Learning Representations}} (2022)

\bibitem{wu2022graph}
Wu, Z., Trevino, A.E., Wu, E., Swanson, K., Kim, H.J., D’Angio, H.B., Preska, R., Charville, G.W., Dalerba, P.D., Egloff, A.M., et~al.: {Graph deep learning for the characterization of tumour microenvironments from spatial protein profiles in tissue specimens}. Nature Biomedical Engineering pp. 1--14 (2022)

\end{thebibliography}
